\documentclass[runningheads]{llncs}
\usepackage{graphicx}
\begin{document}
\title{Usage-based learning of grammatical categories}
\titlerunning{Usage-based learning of grammatical categories}
%
\author{Luc Steels\inst{2} \and Paul Van Eecke\inst{1} \and Katrien Beuls\inst{1}}
\authorrunning{L. Steels et al.}
\institute{VUB Artificial Intelligence Laboratory, 
Pleinlaan 2, 1050 Brussels, Belgium \and
ICREA, Instituci\'{o} Catalana de Recerca i Estudis Avan\c{c}ats,\\
Institut de Biologia Evolutiva (UPF-CSIC)\\
Dr. Aiguadar 66, Barcelona, Spain, 08003\\
}
%
%
%
\maketitle              
\noindent
\textit{Published after double-blind review as:\\
Steels, L., Van Eecke, P. \& Beuls, K. (2018). Usage-based learning of grammatical categories.Belgian/Netherlands Artificial Intelligence Conference (BNAIC) 2018 Preproceedings (pp. 253-264).}

\begin{abstract}
Human languages use a wide range of grammatical categories to constrain which words or phrases can fill certain slots in grammatical patterns and to express additional meanings, such as tense or aspect, through morpho-syntactic means. These grammatical categories, which are most often language-specific and changing over time, are difficult to define and learn. This paper raises the question how these categories can be acquired and where they have come from. We explore a usage-based approach. This means that categories and grammatical constructions are selected and aligned by their success in language interactions. We report on a multi-agent experiment in which agents are endowed with mechanisms for understanding and producing utterances as well as mechanisms for expanding their inventories using a meta-level learning process based on pro- and anti-unification. We show that a categorial type network which has scores based on the success in a language interaction leads to the spontaneous formation of grammatical categories in tandem with the formation of grammatical patterns. 
\end{abstract}

\section{The nature of grammatical categories}

It is well known that human languages mediate the bi-directional mapping between meaning and form 
through a very large set of grammatical constructions and a complex web of grammatical categories 
used by these constructions. A typical example is the ordering of adjectives before the noun 
in English noun phrases, as in ``a beautiful old red Italian Vespa bike''. These adjectives 
follow a preferred order, known as the `royal order of adjectives', with the sequence of
slot fillers often described in terms of categories such as observational, 
physical (with size - shape - age - color), origin, material, and qualifier. An utterance, such as 
``an Italian red old Vespa beautiful bike'', would perhaps still be understood but is experienced by 
native speakers as non-standard. The problem is that there are many other types of adjectives not 
covered by these categories and also that it is not always easy to categorize adjectives in 
terms of these five main categories.

Human languages use certainly hundreds, perhaps thousands of grammatical categories. 
They can be organised in categorial hierarchies (similar to type hierarchies). For example, 
human and animal are both subcategories of animate; article, possessive, number and demonstrative are 
usually considered subcategories of determiner. 
Some categories are used in agreement systems (such as subject-verb agreement). Others express 
additional aspects of meaning 
through morphosyntactic means, such as tense (present/past/future), or definiteness (definite/indefinite). 
Here we focus on categories that constrain which slots certain words or phrases can fill in a 
grammatical construction. They may
reflect semantic aspects (such as animate or shape) or the role of a class of words or 
phrases in the grammar (such as determiner or modal verb). 

Grammatical categories are difficult to define. They are 
often grounded in cultural habits of the linguistic community \cite{Talmy:2000}, almost never without 
exceptions \cite{Ramat:1999}, and divergent from one language to another \cite{Haspelmath:2007}.
Nevertheless a lot of work in linguistics as well as machine learning 
assumes that grammatical categories are static and language-independent, and subsequently 
that a sufficiently big corpus of frozen language use annotated with a set of universal 
grammatical categories, such as the proposed universal dependencies \cite{Marneffe:2014} or 
universal part-of-speech tags \cite{Petrov:2012}, is effective for learning any language. 

However, according to another view (expressed for example by \cite{Hopper:1991}), grammatical 
categories in language diverge between languages and a language is forever 
emergent: there are ongoing innovations or shifts in possible meanings and functions of 
words, or grammatical structures. These changes come up through language use and are 
not always fully conventionalized. Hence language becomes viewed as a complex adaptive 
system that is always on the move \cite{Steels:2000}. 
If grammars and lexicons are emergent and changing, grammatical categories also will be emergent, 
language-specific, diverse across individuals and languages,  and forever changing as the grammatical constructions in which they function change.

The present paper explores this 'emergent' view on language by investigating how a population of 
agents could come up with a grammar, and therefore a set of grammatical categories,
without prior design and without a prior corpus of human language use. 
Instead the agents have to invent the grammar from scratch, without 
a central agency or telepathy. This ambitious goal 
pushes the exploration of emergent language to a limit where we can address how (new) 
grammatical categories form and spread. 

\section{The Syntax Game}

We have set up an experiment in which a 
population of agents plays language games \cite{SteelsKaplan:1999}
and constructs a language as part of the game. We want a game with minimal complexity but 
enough challenges to explore the formation of grammatical categories. 
So we have used the Syntax Game, a referential game that is similar 
to the well known and extensively studied Naming Game \cite{SteelsKaplan:2002}, and has 
already been used in earlier experiments on the emergence of hierarchies 
and recursion \cite{SteelsWellens:2006}, \cite{deBeule:2008}, \cite{SteelsBleys:2007}, 
\cite{SteelsGarcia:2015}. In the classical Naming Game, the topic 
is a single object to which the speaker wants to draw the attention, the meaning is a 
single category, e.g. a color, and the utterance a single word. The Syntax Game allows topics 
to be a set of objects, with the number of objects variable and not known to the listener; 
meanings are a conjunction of categories where categories can be both properties (unary 
relations) as well as relations with multiple arguments; and the utterance consists 
of multiple words, put together according to the syntactic conventions emerging within 
the population. A word can introduce one or more predicates. 

\begin{figure}[h!]
\centering
\includegraphics[width=.5\textwidth]{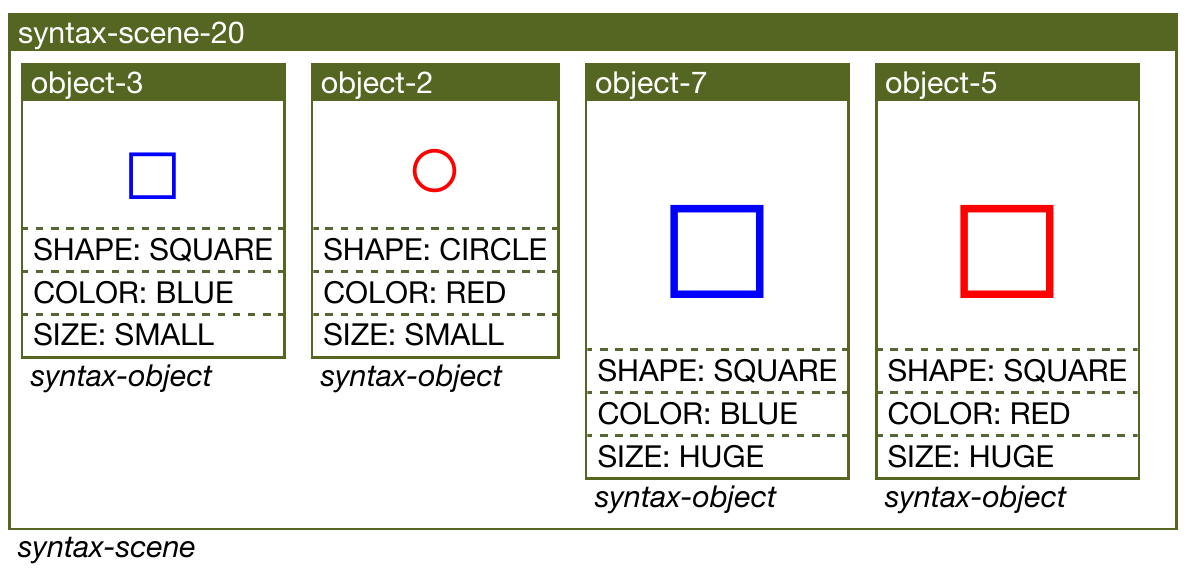}\\
\includegraphics[width=.5\textwidth]{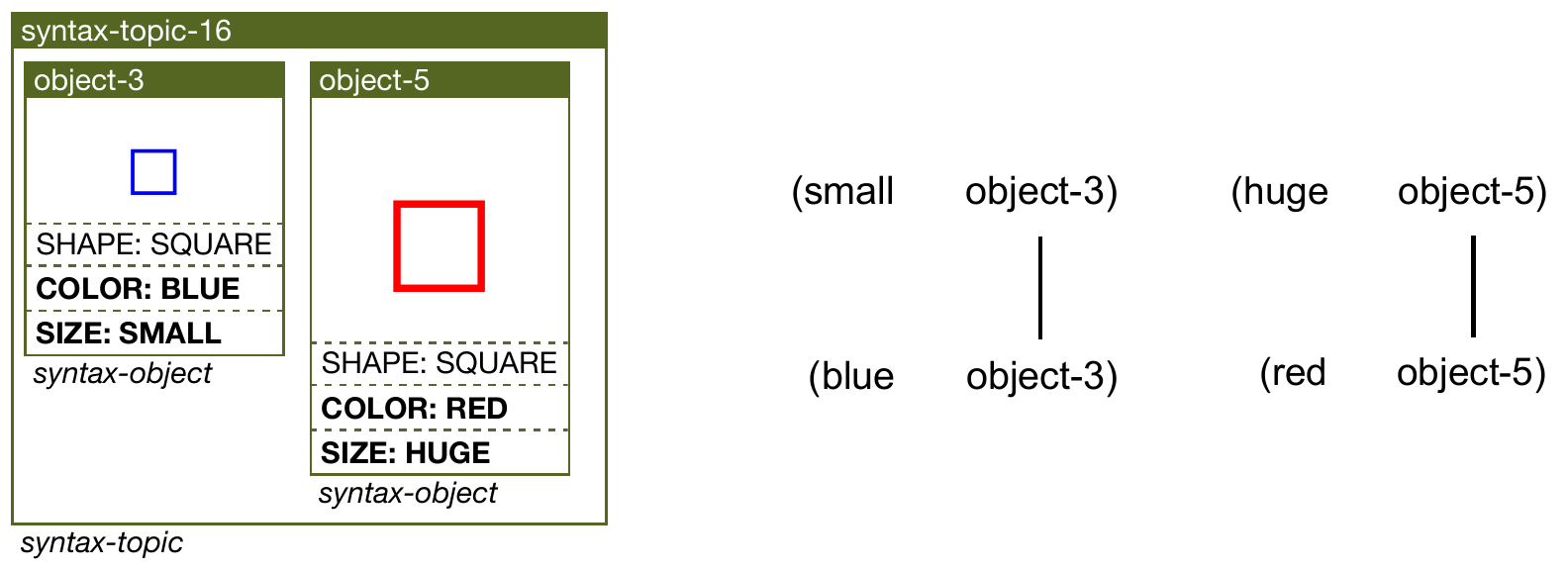}
\caption{{\it Top:} An example scene of four objects in the form of geometric figures with dimensions 
for shape, color, and size. {\it Bottom left:} The topic of the utterance consists of two objects chosen 
from this scene. {\it Bottom right:} The meaning to be expressed by the speaker consists of 
a combination of features discriminating these objects from the others in the context.}
\label{fig:syntax-scene}
\end{figure}

In the experiments reported later, the environment consists of geometric figures 
with different colors, sizes and shapes. A specific scene, which is the context for 
a game, consists of a set of figures (see Figure \ref{fig:syntax-scene}) and in each game the speaker 
chooses one or more objects from the scene as the topic. 

Without grammar, the Syntax Game quickly generates a combinatorial search problem for 
semantic interpretation \cite{SteelsGarcia:2015} because 
the listener cannot know which words cooperate for referring to a particular object, which 
objects fill the different arguments of relations, and how many objects are being referred to. 
For example, assuming there is no grammar, the listener cannot know - without access to the 
scene - whether the utterance ``big brown red small left-of triangle square''
refers to a single object, namely `a big, brown, triangle, which is left-of a red small square', or 
to two objects: `a red triangle left of something big' and `a brown square', or to a
bigger collection of objects. In fact, without grammar,
the set of possibilities is equal to all possible ways to partition the set of words in the 
utterance. The cardinality of this set is equal to the Bell number known to be 
double exponential \cite{SteelsGarcia:2015}.

The use of grammar alleviates this problem drastically by establishing co-reference links between 
the arguments of the various predicates introduced by these words. For example, suppose there 
is an English-like grammar and the speaker produces the 
utterance ``(a) big brown triangle left of (a) red small square''. When the listener 
performs lexicon lookup only, the following set of predicates is obtained: 
\begin{quotation}
[1] big(?x1), brown(?x2), triangle(?x3), red(?x4), \\ small(?x5), square(?x6), left-of (?x7,?x8)
\end{quotation}
However when taking into account the grammar, [1] can be transformed into [2]: 
\begin{quotation}
[2] (topic ?x) big(?x), brown(?x), triangle(?x), \\ red(?y), small(?y), square(?y), left-of (?x,?y)
\end{quotation}
The listener now knows that there are only two 
objects involved, which predicates are relevant for referring to which one, and which object is 
the topic. Concretely, he knows example that the predicates `big', `brown', and `triangle' have 
the same argument because they are all part of the same noun phrase. So the task of the grammar is to 
transform [1] into [2] and thus dampen the combinatorial explosion in semantic interpretation. 

\section{Fluid Construction Grammar}

\begin{figure}
\centering
\includegraphics[width=1.0\linewidth,height=.9\textheight,keepaspectratio]{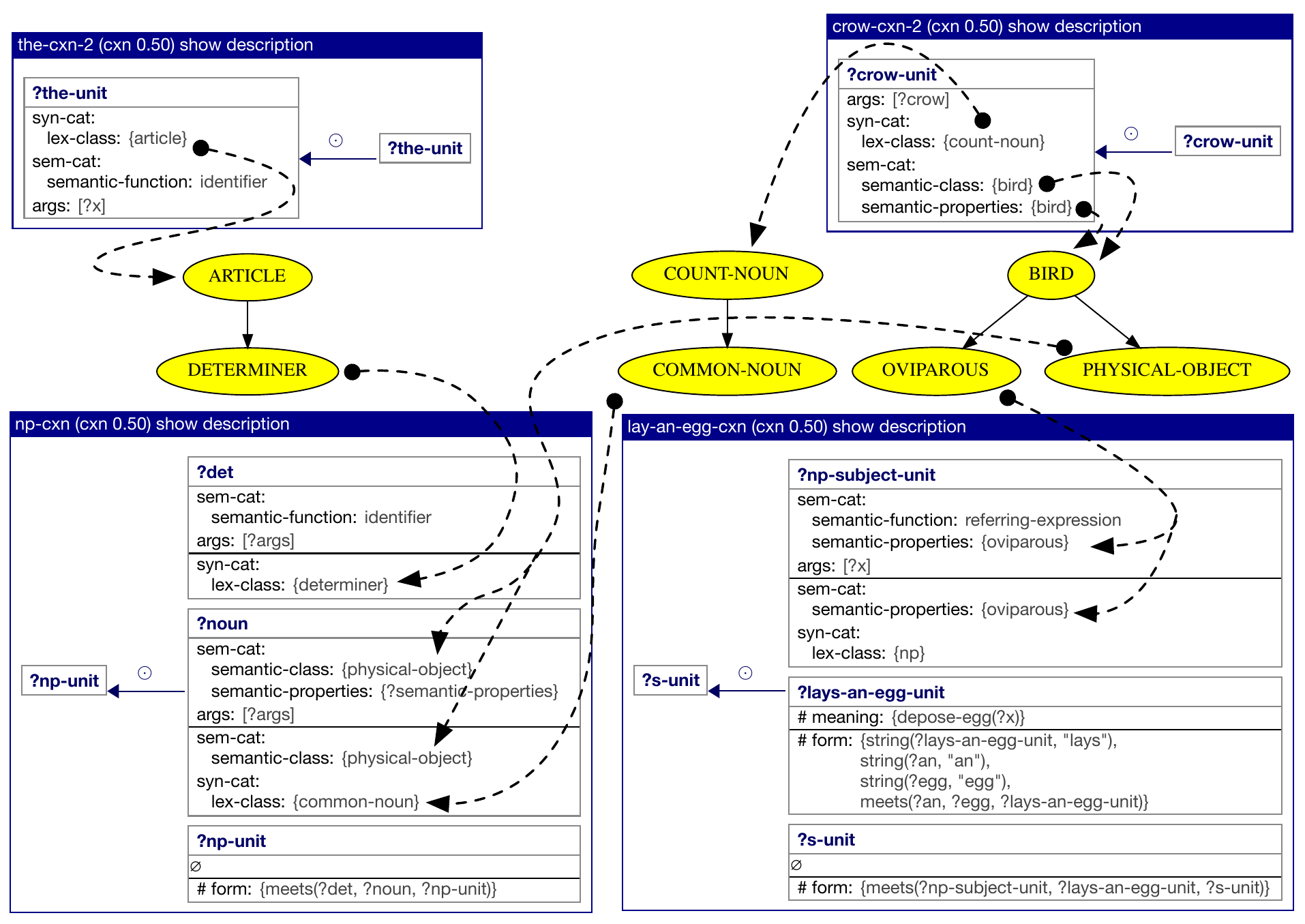}
\caption{Construction schemas and categorial networks for processing ``the crow lays 
an egg' in Fluid Construction Grammar. From the left top clockwise, there is a construction for
the word ``the", the word ``crow", a noun-phrase construction and an item-based construction, 
typical for early stages of grammar \cite{Tomasello:2005}. 
The arrows show the categories that are matched and merged via the categorial network.}
\label{fig:diversity-2}
\end{figure}

We have used Fluid Construction Grammar (FCG) \cite{Steels:2013} as basic formalism in the experiments.
It supports grammars in the form of construction schemas, i.e. associations between semantic constraints
on meaning and syntactic constraints on form. Each construction schema has a set of unit-slots with unique 
names and associated features and values (see Figure \ref{fig:diversity-2}). 
FCG follows the tradition of feature-structure based formalisms, such as HPSG, in the 
sense that the main data structure is a feature structure both for construction schemas and for 
the transient structure which contains information about the utterance being built or parsed. 
FCG uses also a form of unification for {\it matching} constructions 
against the transient structure in order to determine whether they apply, and for {\it merging} 
information from the construction into the transient structure if this is the case. 
Both semantic parsing and sentence production proceed by the 
successive application of construction schemas. The unavoidably search is handled 
with heuristics and priming \cite{Steels:2012}.

Importantly, for the present paper, matching and merging makes use of a
{\it categorial network} which defines for each category, 
for example `count-noun', whether it can be a subcategory of another one, for 
example `common-noun' (see Figure \ref{fig:diversity-2}). Thus, 
when the category `count-noun', which is the lexical class of ``crow", is matched with the 
category `common-noun', for being the lexical class of the head 
of a noun-phrase, direct unification fails. However, because there is a path between `count-noun' and 
`common-noun' in the categorial network (as illustrated in Figure \ref{fig:diversity-2}), they unify. 
So the main role of the categorial network is to make the matching and 
merging process more powerful by exploiting subcategory relations. This avoids 
endless multiplication of construction schemas and increases the generalization 
power of the grammar greatly. 

\section{Usage-based learning of construction schemas} 

There are many approaches to category formation and grammar learning currently being explored. 
Most of them perform some form of statistical learning based on data of a very large corpus of 
language usage. This is certainly a fruitful avenue of research. But here we are interested 
in a situation where there is no prior corpus and agents have to come up with their own constructions
and the categories that go with them. This poses two problems (i) how are new construction schemas
built, and (ii) how does the population arrive at a sufficiently shared grammar to 
guarantee successful communication. 

Usage-based learning tackles the first problem by a {\it constructivist} approach. 
Speaker and listener try to handle each language game first using their existing inventory of 
lexico-grammatical construction schemas. However when the existing schemas do not 
work, they create new construction schemas: 
\begin{enumerate}
\item The speaker can find out whether there are missing construction schemas by {\it re-entrance}: 
He simulates the parsing process of the listener and when there 
is ambiguity or combinatorial uncertainty in semantic interpretation, he 
creates a construction that introduces 
syntactic constraints (i.c. grouping and word order) to signal the missing 
semantic information to the listener. 
For example, if the speaker wants to express the 
conjunction big(?x), triangle(?x), and uses 
the utterance ``big block" or ``block big" (no grammar assumed), then he will detect after re-entering 
and parsing this utterance that it does not convey that the arguments of the
predicates `big' and `block' are the same. The speaker then builds a new construction schema with 
a unit that combines these words into a new unit and makes the variables equal. Later on 
he can reuse this construction schema in the same situation. 
\item The hearer uses his own lexicon and grammar in a flexible way to parse the utterance as well as 
possible and tries to get the best possible interpretation, after inspecting the scene 
and getting feedback from the listener. After semantic parsing and interpretation, 
the hearer can then detect missing, only partially matching, or incomplete construction schemas
that would allow parsing and interpretation, and he 
subsequently adds these to his own inventory. For example, he interprets the 
sequential order of the words as a cue that the variables of the two predicates are co-referential. 
\end{enumerate}
New construction schemas are often built from existing construction schemas which almost match, 
so that there is a gradual generalization and specialization of construction schemas. 
This has been implemented in the experiments reported here 
using a combination of anti-unification and pro-unification. Anti-unification is familiar 
from constraint logics and inductive logic programming \cite{Page:1993}. 
Whereas unification computes the most general specialisation of two feature structures, 
anti-unification computes the least general generalisation. This can involve making constants 
variable, splitting the same variable into different variables, relaxing constraints, 
ignoring the presence of units, etc. Anti-unification gives information for the learner how 
to generalise an existing construction schema to make it fit with a novel situation. To avoid 
overgeneralization, pro-unification then takes a construction schema generalized 
through anti-unification and constrains it again based on properties of the 
current situation so that it is more specific \cite{SteelsVanEecke:2018}.

This constructivist step generates possible construction schemas but there is no guarantee that 
these are the ones adopted by the rest of the population. For example, the speaker could have 
made a construction schema combining ``red" and ``triangle" into a unit ``red triangle" in order 
to signal that both words are introducing predicates about the same object, not knowing that 
other agents had already made a construction schema where ``red" follows ``triangle" to make the 
 unit ``triangle red". The second problem, namely how construction schemas get shared in the 
population, is solved using a {\it selectionist} lateral inhibition familiar from the Naming Game. 
It ensures that those construction schemas which are effective in language games survive 
and that the grammars of different agents get aligned without a central coordinator. 

The selectionist step is implemented by associating a score with each construction schema 
to determine which schema is to be preferred when performing search.
\begin{enumerate}
\item If a construction schema {\it s} is used in a successful language game, then the score of {\it s}
is increased and the scores of its competitors decreased. Competitors are other construction 
schemas that could also trigger but were either not pursued further because {\it s} was heuristically 
preferred in search, or the competitor was pursued but it lead to a dead-end with subsequent 
backtracking to {\it s}. 
\item If a construction schema {\it s} was used in a failing language game, then the score of {\it s} is 
decreased. 
\end{enumerate}
Even without using a categorial network, populations can be seen to converge towards a 
common set of construction schemas that is able to play the language game 
successfully \cite{SteelsGarcia:2015}.

\section{Usage-based learning of grammatical categories}

The key novelty of the present paper lies in the way subcategory links are acquired. 
We focus on grammatical categories that 
constrain what feature values the units filling a particular slot in a construction schema need to have. 
Grammarians use semantic or syntactic ways to characterize these slots but we hypothesize here 
that categories emerge based on usage in the emergent grammar. The fact that they 
reflect semantic or syntactic properties is a side effect of the structure of the domain, 
the relevance of these properties in language games, and the emergent grammar (for example how many 
slots there are in a phrase). This would explain why grammatical categories are so difficult 
to define and learn. As with construction schemas, there are two issues: how categories are 
acquired and how categories become shared and aligned in the population. 

Categories are acquired using a constructivist approach based on need. Suppose there is a 
feature value {\it v} 
for feature {\it f} of a particular unit in a construction schema and it is not matching 
with a corresponding feature value {\it v'} for the same feature {\it f} in a corresponding unit 
in a transient structure, 
then the agent can build a supercategory {\it c} with subcategory links from {\it v'} and 
{\it v} to {\it c}. The new supercategory now covers both {\it v} and {\it v'} and unification
can succeed through the categorial network. If {\it v} had already a supercategory {\it c} then 
it can be reused by making {\it v'} another subcategory of {\it c}. 

But just as for construction schemas, the learner cannot know whether the appropriate  
subcategorisation links have been added to the categorial network. For example, it is possible 
that the learner introduces subcategory links from article and possessive to determiner, more precisely 
between a supercategory `elements that can fill the determiner
slot in the noun-phrase' to `elements that can fill an article slot or a possessive slot'. But 
another agent might not put possessives in the determiner slot, but in the adjectival slot instead.
There is no right or wrong, only the collective dynamics of the population determines what 
is to be the shared convention. 

We use the same selectionist lateral inhibition approach as for construction schemas. 
Each subcategory relation {\it r(c$_1$,c$_2$)} has an associated score. The score is used to 
decide whether a particular subcategory should be preferred. 
The score is updated based on the outcome of a language game: 
\begin{enumerate}
\item If the subcategory relation {\it r} was used to make matching and merging succeed 
in a successful language game, then its score is increased and the scores of its 
competitors decreased. Competitors are other subcategory relations involving {\it c$_1$}
that were not preferred in the search process, or were used but lead to a dead-end with subsequent 
backtracking. 
\item If a subcategory relation was used in a failing language game, then its score is decreased. 
\end{enumerate}
\begin{figure}
\centering
\includegraphics[width=0.85\textwidth]{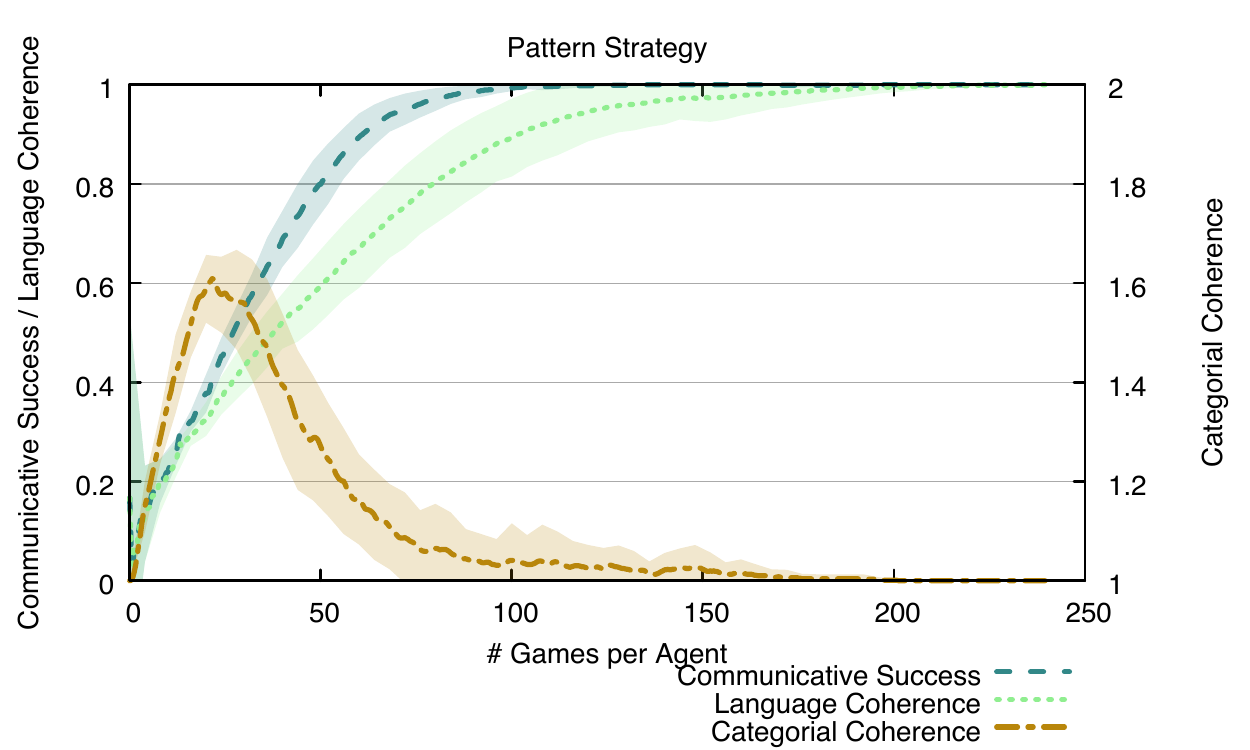}
\caption{Results of a simulation in which ten agents build an emergent grammar. The blue line indicates communicative success and the green line grammatical coherence (left y-axis). The red line indicates
categorial coherence (right y-axis). See text for explanation.}
\label{fig:outcome}
\end{figure}

\section{Experimental results} 

The experiment shown in 
Figure \ref{fig:syntax-scene} uses the geometric figures introduced earlier. There are three dimensions, 
color (yellow/blue/green/red), size (tiny/small/large/huge), and shape (square/triangle/circle), and the target of the agents is to come up with a grammar to be 
successful in the Syntax Game, using a minimal number of shared grammatical construction schemas, 
which is here equal to 4 (assuming that single word utterances do not require grammatical constructions). 
This is only possible when they develop a (shared) categorial network, and the question is what 
this network is going to look like. 

\begin{figure}
\centering
\includegraphics[width=.8\textwidth]{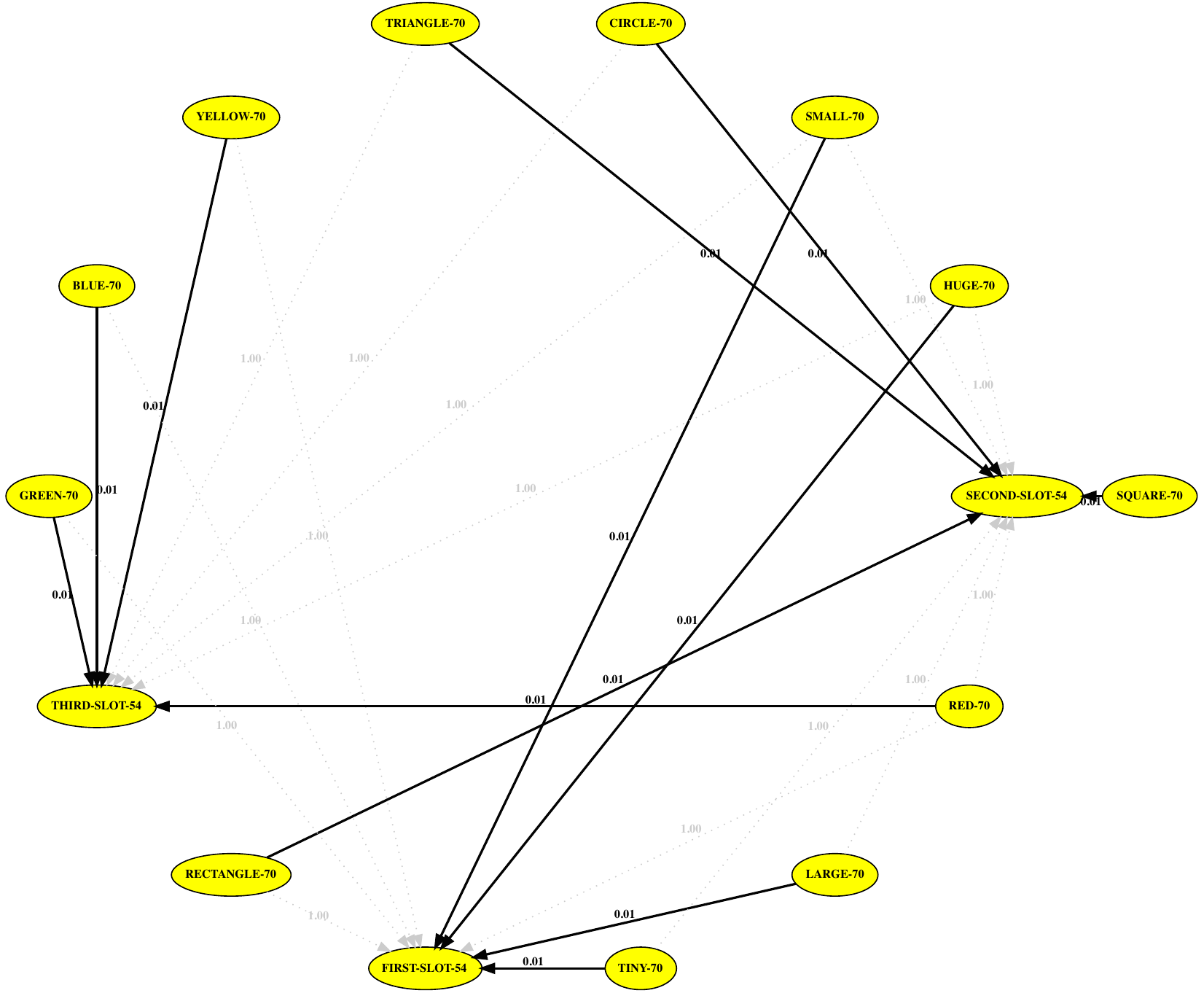}\\
\includegraphics[width=.8\textwidth]{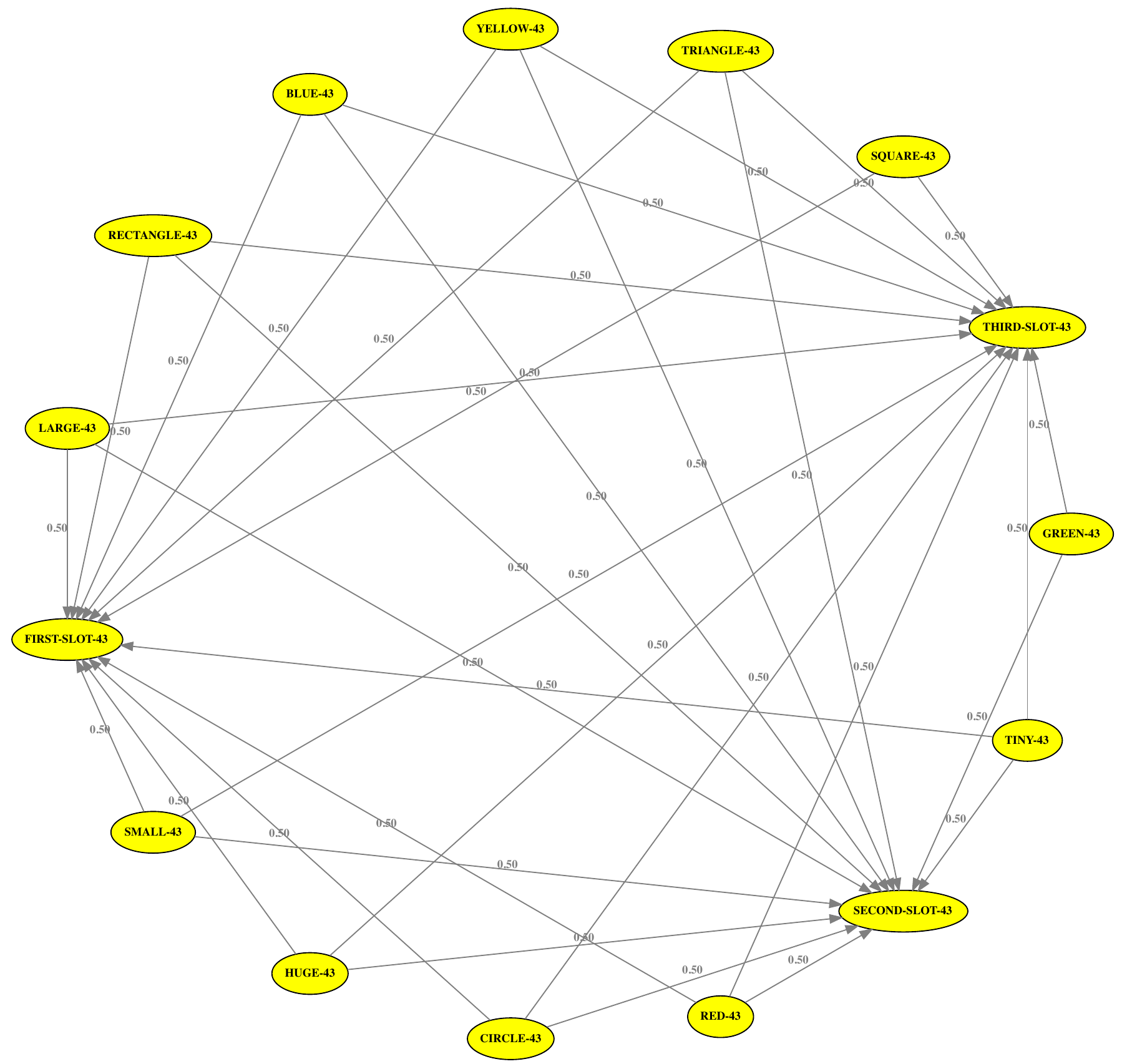}
\caption{{\it Top:} The Categorial Network of a single agent in an experiment which uses 
the selectionist lateral inhibition update of scores. {\it Bottom:} The Categorial Network of a single 
agent without this update rule.}
\label{fig:categorial-network}
\end{figure}

Figure \ref{fig:outcome} shows the average results of a series of 6 experiments 
with 10 agents on average playing 250
games each. Only one game is played by 2 agents per time step. The figure displays communicative success in 
the language game (blue line), which steadily increases to reach 100 \% after 100 games per agent. 
As games proceed, we see that the grammatical coherence (green line) also increases to reach 100 \%. 
Grammatical coherence is calculated based on whether the listener would reproduce the same utterance
as the speaker in a language game. The categorial coherence (orange line, right x-axis) shows that agents 
settle on a single category for each word. Most remarkably there is a correspondence between 
the semantic dimensions and the grammatical categories - even if the agents do not use their 
ontology to come up with grammatical categories. This is shown in 
Figure \ref{fig:categorial-network} where we see first (top)
a network where agents use the adjustment of scores based on usage in language games and 
next (bottom) a network where agents do not employ a usage-based strategy, i.e. the success or failure 
in the language game does not play a role in updating of the sub-categorisation relations. 

We refer the reader to
\cite{VanEecke:2018} for extensive explanations of all steps in the language game, examples of 
grammatical constructions and categorial networks, and more experiments.

\section{Conclusion}

The formation of an emergent grammar includes both the invention and learning of grammatical 
construction schemas and grammatical categories used to define constraints on slot fillers in 
these schemas. This paper explored a usage-based approach, meaning that the categories and 
construction schemas are built in order 
to handle communicative failures, e.g. too much ambiguity or missing construction schemas in 
parsing and production, and once added, they enter into a competition against others, with the 
primary selectionist force being success in 
a language game. Remarkably, we see that the grammatical categories emerging this way 
are to some extent semantically motivated but this is a side effect of the communicative situations
and the nature of the emergent grammar and not due to a prior ontology. 

The results reported here shed light on the origins and acquisition of grammatical categories, 
a notoriously difficult question in language research. Although most language research today
assumes that grammatical categories are given a priori and are universal, the study of a diverse range 
of languages and of rapid linguistic change, particularly on social media, shows that this is an 
idealization. However, the mechanisms discussed here show that this does not mean that the problem 
of grammar emergence and learning is hopeless. 

\section{Acknowledgment}
This paper was funded by the Atlantis Project (CHIST-ERA) (for LS, PVE and KB) and by ICREA and the Institute for Evolutionary Biology (UPF-CSIC) in Barcelona for LS. 

\bibliography{steels2018usage}
\bibliographystyle{splncs04}

\end{document}